\documentclass[conference]{IEEEtran}
\IEEEoverridecommandlockouts
\usepackage{cite}
\usepackage{amsmath,amssymb,amsfonts}
\usepackage{algorithmic}
\usepackage{graphicx}
\usepackage{textcomp}
\usepackage{xcolor}
\usepackage{pifont}
\newcommand{\cmark}{\ding{51}}%
\newcommand{\xmark}{\ding{55}}
\usepackage{color, colortbl}
\usepackage{multirow}

\def\BibTeX{{\rm B\kern-.05em{\sc i\kern-.025em b}\kern-.08em
    T\kern-.1667em\lower.7ex\hbox{E}\kern-.125emX}}
\begin{document}

\title{ 
    PANDA : Perceptually Aware Neural Detection of Anomalies
}
\author{\IEEEauthorblockN{Jack W. Barker $^{1}$, Toby P. Breckon $^{1, 2}$}
\IEEEauthorblockA{\textit{Department of Computer Science$^{1}$ $|$ Department of Engineering ${^2}$} \\
\textit{Durham University, UK}\\
}

}

\maketitle

\begin{abstract}
  Semi-supervised methods of anomaly detection have seen substantial advancement in recent years. Of particular interest are applications of such methods to diverse, real-world anomaly detection problems where anomalous variations can vary from the visually obvious to the very subtle. In this work, we propose a novel fine-grained VAE-GAN architecture trained in a semi-supervised manner in order to detect both visually distinct and subtle anomalies. With the use of a residually connected dual-feature extractor, a fine-grained discriminator and a perceptual loss function, we are able to detect subtle, low inter-class (anomaly vs. normal) variant anomalies with greater detection capability and smaller margins of deviation in AUC value during inference compared to prior work whilst also remaining time-efficient during inference. We achieve state-of-the-art anomaly detection results when compared extensively with prior semi-supervised approaches across a multitude of anomaly detection benchmark tasks including trivial leave-one-out tasks (CIFAR-10 - AUPRC$_{avg}$: 0.91; MNIST - AUPRC$_{avg}$: 0.90) in addition to challenging real-world anomaly detection tasks (plant leaf disease - AUC: 0.776; threat item X-ray - AUC: 0.51), video frame-level anomaly detection (UCSDPed1 - AUC: 0.95) and high frequency texture with object anomalous defect detection (MVTEC - AUC$_{avg}$: 0.83).

\end{abstract}

\section{Introduction}
Anomaly detection is the task of recognising samples of a given dataset which deviate significantly from established normality and as such, represent unexpected eventualities or outliers in the scope of a given task \cite{akcay2020}. Anomaly detection is a challenging task because of the broad range of variational forms which anomalies may present, representing an unbounded (open set) distribution of possible deviations from normality.

The ultimate application challenge of anomaly detection models is effective and efficient applications to critical real-world tasks \cite{Narayan2020, Hoque2020, Fang2020} with little to no human intervention at training time. However, by contrast, the \textit{modus operandi} of anomaly detection evaluation in the literature is to demonstrate model performance across trivial and unrealistic ‘leave one out’ tasks on general datasets such as MNIST \cite{lecun-mnisthandwrittendigit-2010} or CIFAR-10 \cite{Krizhevsky09learningmultiple} in which one class from the dataset is labelled as anomalous and all other classes as normal. This evaluation methodology is highly unrealistic as not only are these datasets not intended for anomaly detection, but the act of directly comparing the classes present in the datasets in this way is unlikely to occur in a real-world anomaly detection application. Anomalies occurring within real-world problems may be subtle, localised to a small sub-region of the image, exhibit high variance and even be the result of subterfuge by an adversary \cite{Hughes2015, Bhowmik2019, Mahadevan2010, Bergmann2019MVTEC}.

Anomaly detection methods have had varying degrees of success across real-world tasks including, but not limited to: retinal diagnosis \cite{Schlegl2017, Schlegl2019}, airport security scanning \cite{Gaus2019, Bhowmik2019, Akcay2018, Akcay2019} and factory line inspection \cite{Bergmann2019MVTEC}. However, these methods can often attribute their limited success to being domain-specific and are not applied across multiple, diverse (multi-spectral; cross-domain) datasets. Such prior anomaly detection methods are overly focused on more general features to aid in categorising visually obvious anomalies \cite{Akcay2018, Schlegl2017} akin to the flawed evaluation methodology of `leave-one-out tasks', meaning they do not perform overly well with visually subtle anomalies in tasks such as \cite{Hughes2015, Schlegl2017, Schlegl2019}.

\begin{figure}[t!]
\centering
\includegraphics[width=250pt]{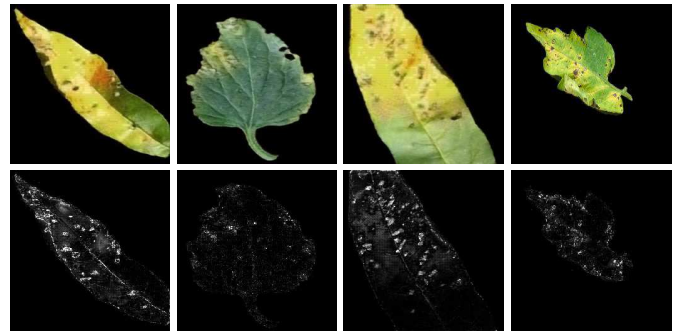}
\caption{\textbf{Top:} Leaves from Plant Village \cite{Hughes2015} featuring visible diseases. \textbf{Bottom:} Anomalous instance segmentation masks generated by PANDA for the respective diseased leaves.}
\label{fig:plant_images_mask}
\end{figure}

Whilst supervised methods \cite{Gaus2019, Bhowmik2019} can obtain superior performance across anomaly detection benchmark tasks, often by following a simplistic anomaly detection by classification paradigm with discrete classes, they require large, labeled-datasets for training. These can be both expensive to obtain, unbalanced in nature, and will always struggle to provide sufficient coverage of rare, low-occurrence anomalies given the potential open-ended scope of the anomalous class space. These challenges of training data adequacy could lead to potential adversarial example attacks against such methods \cite{Goodfellow2015, Paudice2018}.

By contrast, semi-supervised methods \cite{Schlegl2019, Baur2018, Vu2019, Akcay2018, Akcay2019} overcome this issue by learning a close approximation to the true distribution manifold exclusively over the non-anomalous (normal) data samples \cite{Schlegl2017}. Such techniques use generative methods in order to approximate the distribution \cite{Akcay2018, Akcay2019}. Furthermore, methods such as \cite{Schlegl2017, Schlegl2019, Akcay2019} suffer from slow inference times which can hinder real-world applicability in scenarios where high-throughput processing is required. Methods such as  \cite{Akcay2018, Schlegl2019} exhibit vastly differing accuracy with each iteration over the same dataset leading to a sparse confidence interval as demonstrated in our experiments, impeding real-world applicability due to unpredictable detection behaviour at inference.

In this work, we propose the Perceptually Aware Neural Detection of Anomalies (PANDA-GAN), a Variational Autoencoder Generative Adversarial Network (VAE-GAN) based architecture to combat the task of detecting subtle fine-grained anomalies present in real-world anomaly detection applications whilst also retaining time-efficiency at inference. PANDA includes three novel proposals: (1) a Fine-Grained Visual Categorisation Discriminator Network (FGVC) to ease the problem of detecting visually subtle, low inter-class variance anomalies present in anomaly detection problems and to provide a harsher critic during training for the GAN generator module; (2) a residually connected dual-feature extractor implementation within our generator module that carries lower-level features in given images forward and combines them residually with higher-level, later features in the architecture; (3) a perceptual loss function based on feature error instead of raw pixel-error \cite{Johnson2016} that in turn obtains higher-fidelity images, but has not yet been applied to the task of generative anomaly detection. This work represents the first instance of these techniques being jointly applied to semi-supervised anomaly detection.

\section{Related Work}

Many works have addressed the problem of generative semi-supervised anomaly detection. Initial methods propose the use of Variational Autoencoder (VAE) architectures \cite{Kingma2013}, in which a latent representation $z$ is learned from the image space $X$ though the use of an encoder. A second module (decoder) then maps from $z$, back to the image space to produce $\hat{X}$. The encoder and decoder can be trained using the reconstruction error between the original image $x\in X$ and the produced image $\hat{x} \in \hat{X}$. Early implementations of VAE \cite{Kingma2013}, however do not capture the distribution of the data $p(X)$ well due to the over-simplification of the learned prior probability $p(z|X)$. VAE are only capable of learning a uni-modal distribution, which fails to capture complex distributions that are commonplace in real world anomaly detection scenarios.

Generative Adversarial Networks (GAN), first proposed by Goodfellow \textit{et al.} \cite{Goodfellow2014}, combat this simplification by forcing a Generator, $G$, to model complex distributions in data from random noise in order to generate representative image samples drawn from this distribution. The learning objective of $G$ is to reduce the confidence of the Discriminator, $D$, to assign an effective probability to a presented image; whether or not it is original or a generated image. The zero-sum end-game of a GAN is the Nash Equilibrium whereby the Generator network and Discriminator network have saturated learning such that the probability of the Discriminator to distinguish between real and synthetically generated images converges. 

AnoGAN \cite{Schlegl2017} is the first GAN-based, semi-supervised method of anomaly detection. The model is trained only on non-anomalous data to learn the manifold $z$ of normal samples. When an anomaly $x_{a}$ is processed by the Generator network, it produces a non-anomalous image $x_{a}'$. Taking an $l_{2}$ reconstruction error will outline anomalies present in the image. Although this method proved that it was possible to use GAN for the task of anomaly detection, the computational performance is incredibly slow hence limiting real world applicability. GANomaly \cite{Akcay2018} overcomes these issues by training a Generator network and a secondary encoder in order to map the generated samples into a second latent space $\hat{z}$ which is then used to better learn the original latent priors $z$, mapping between latent values efficiently at the same time as the Generator learns the distribution manifold over data $x$. Follow-on work, Skip-GANomaly\cite{Akcay2019}, introduces the notion of skip connections into the network architecture in order to preserve image information detail across the encoder-decoder structure, greatly improving performance. 

Concurrently, the author of AnoGAN produced the Fast-AnnoGAN \cite{Schlegl2019} method of anomaly detection which is similar to their previous work, but replaces the Deep Convolutional GAN (DCGAN) \cite{Radford2016} with a Wasserstein Generative model \cite{Arjovsky2017}. It uses the trained Generator from the GAN training to train an encoder which maps images to the latent space. This enables the overall Fast-AnnoGAN architecture to avoid the computationally expensive operation of obtaining a latent representation at inference. A similar approach was adopted by Houssam \textit{et al.} \cite{Zenati2018} which is based on the BiGAN architecture \cite{Donahue2019}. They use an approach similar to the one indicated in \cite{Donahue2019} to solve the optimisation problem $min_{G,E} max_{D}V(D,G,E) $ where the features of $X$ are learned by the network $E$ to produce the pair $(x, E(x))$. The Generator network learns the pair of $(z, G(z))$ from the real features of $X$ \cite{DiMattia2019}. This simultaneous learning of the pairs forces the network to learn the mapping from not only image data to latent space, but from latent space back to image data. 
 
Although GANs have risen to prominence and gained significant results, producing high-fidelity images, they suffer from volatile training issues such as mode collapse and instability, leading to over-fitting during training. VAE-GAN \cite{Larsen2016} are VAE which are trained in an adversarial manner. The first instance of these being utilised for anomaly detection is Baur \textit{et al.} \cite{Baur2018} which utilises a common VAE architecture, but applies a Discriminator network in order to determine whether the images are real, or reconstructed. Similar work was performed in ADAE \cite{Vu2019} which uses a dual, parallel model in which the primary VAE is the Generator and the second VAE is the Discriminator. As well as reconstructing the images from the latent representation, the authors also compute the error between the data distributions between the dual networks in order to gain a closer fit for the manifold over real examples.

\begin{figure*}[t!]
\centering
\includegraphics[width=300pt]{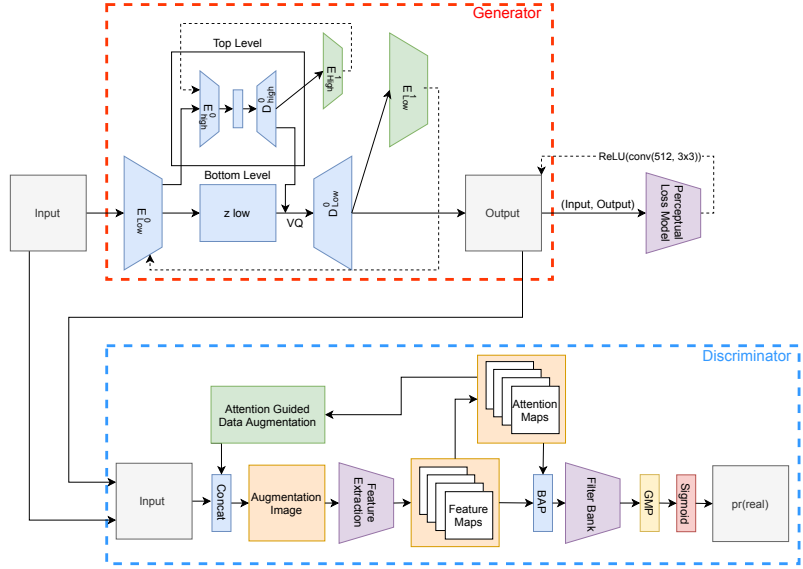}
\caption[caption]{Proposed model architecture featuring our PANDA-GAN architecture with the Generator network (upper) and the Discriminator network (lower).}
\label{fig:model}
\end{figure*}

\section{Approach}

Our proposed method applies a unique VAE-GAN architecture to the task of anomaly detection. Our Generator network (Section \ref{Sec:Generator}) uses skip-connected dual-feature extractor encoders ($\{E^{0}_{high}, E^{0}_{low}\}$) to extract both a top and bottom level latent space which residually combines early, low-level information with high-level features further in the architecture. This network is trained adversarially with a fine-grained Discriminator network (Section \ref{Sec:Discriminator}) which is optimised to assign a true probability that a presented image conforms to the distribution of normality and is not synthetically generated and used to detect subtle discriminating features from input images during inference. A visual overview of our approach is shown in Figure \ref{fig:model}.

\subsection{Generator Network} \label{Sec:Generator}

Our Generator VAE model is trained adversarially to create a VAE-GAN which does not exhibit mode collapse or vanishing gradient training difficulties present in traditional GAN architectures \cite{Goodfellow2014}. Similar to a U-net model architecture, input images $x$ are first fed into a low-level encoder E$^{0}_{low}$ which maps to a latent representation $z_{low}$ = $p_{\theta}(z_{low}|x) \sim N(\bar{x}, \sigma)$. We then further encode $z_{low}$ into a higher-order latent representation $z_{high} = p_{\theta}(z_{high}|z_{low})\sim N(\bar{z_{low}}, \sigma) \cdot p_{\theta}(z_{low}|x)$ which captures higher-level image features. Latent representation $p_{\theta}(z_{low}|x)$ is then combined residually through skip-connections to $p_{\theta}(z_{high}|z_{low})$; meaning that low-level information is preserved upon decoding back into image space $x'$ via $q_{\phi}(x'|z)$. We use only one higher-order latent representation due to memory constraints and to keep our method more efficient during inference. We show that our method obtains state-of-the-art performance by utilising just one higher-order latent representation.

Additionally, we also utilise two secondary Encoders, $\{E^{1}_{high}, E^{1}_{low}\}$ during training exclusively, to re-encode the decoded respective latent representations $\{q_{\phi}(z_{high}|x)|z_{high}), q_{\phi}(z_{low}|x)\} $ back into latent space $z'_{high}$ and $z'_{low}$ respectively. This better models and constrains the latent priors of $\{E^{0}_{high}, E^{0}_{low}\}$ to not be entirely dependent on the input image. This approach was employed in GANomaly \cite{Akcay2018} and Skip-GANomaly \cite{Akcay2019} to yield better performance. $\{E^{1}_{high}$ and $E^{1}_{low}\}$ are not enabled during inference to increase network efficiency.

Overall our VAE-GAN learning objective seeks to minimise over $\forall x \in X$:
\begin{equation}\label{loss_vae}
L_{VAE} =  L_{rec} + C(x, x') + L_{z[0]} + L_{z[1]}
\end{equation}
where $L_{rec}=\left\|x-x'\right\|_{2}$ and $x'$ is generated from the Generator Network: $D^{0}_{low}(D^{0}_{high}(E^{0}_{high}(E^{0}_{low}(x)))\cdot E^{0}_{low}(x))$. C is the critic, or commonly named Discriminator Network (Section \ref{Sec:Discriminator}). Latent loss $L_{z[i]} = \left\|E^{1}_{[i]}-E^{0}_{[i]}\right\|_2 , i = \{high, low\}$.

We also introduce the notion of perceptual loss (PL) to calculate feature error rather than pixel-wise error. Previously introduced to Style Transfer \textit{et al.} \cite{Johnson2016} we introduce usage into the problem of Anomaly Detection instead of using conventional Pixel-Wise Loss (PWL). While PWL computes pixel differences between $x$ and $x'$ on low-level pixel information, PL takes the advantage of taking the error between the high-level activation features \cite{Johnson2016}. Feeding the pair $(x_{i}, x_{i}'), \forall x_{i} \in X$ through a pre-trained conventional Convolutional Neural Network ($f()$) obtains differing activations ($f(x_{i}), f(x_{i}')$) of a given convolutional feature extraction layer. PL is then calculated as $\left\| f(x_{i}), f(x_{i}') \right\|_{2}$. We use a VGG19 \cite{Simonyan15} network as our Perceptual Loss model and take the error between the activations of the 14$^{th}$ layer. We utilise two variants on perceptual loss: 
\begin{itemize}
    \item General perceptual loss (PL$_{g}$()) : Weights obtained by pre-training a CNN trained across ImageNet \cite{Deng2009}.
    \item Problem-specific perceptual loss (PL$_{ps}$()) : Weights obtained from CNN which was pre-trained in a self-supervised fashion over non-anomalous samples from the specific anomaly detection task dataset.
\end{itemize}

\subsection{Discriminator Network} \label{Sec:Discriminator}

In contrast to prior work in anomaly detection which utilise conventional Discriminators \cite{Razavi2019, Schlegl2017, Schlegl2019, Akcay2018, Akcay2019}, in this work we incorporate a Fine-grained Visual Categorisation (FGVC) Discriminator. In general, FGVC is for use in obtaining specific sub-class classification of objects (e.g. species of bird or model of car) \cite{Wang2019}. Typical FGVC datasets are inherently difficult to classify due to highly localised and visually subtle distinguishing features between classes. Within real-world anomaly detection problems, there exist varying levels of anomaly ranging from visually obvious to negligibly subtle. Our FGVC discriminator is optimised to detect these subtle anomalies during inference by recognising the discriminating regions within presented images. It also acts as a harsher critic to our Generator module during training, promoting emphasis on generation of the objects themselves rather than the background context within images.

Our discriminator is inspired from the Weakly Supervised Data Augmentation Network (WS-DAN) architecture \cite{Hu2019}, a proven method in FGVC which obtains superior categorisation performance in the task of FGVC \cite{Hu2019}.

The WS-DAN architecture contains attention layers which allow the network to focus upon both detailed features and key discriminative object parts during inference when categorising anomalous data. This mechanism also allows attention guided data augmentation within the network leading to higher information gain and optimised augmentation of non-anomalous samples during training. The resulting attention maps are combined with feature representations via Bilinear Attention Pooling (BAP). This combined feature representation is then fed into a discriminative filter bank of $1\times1$ convolutions followed by a Global Max Pooling (GMP) \cite{wang2018} layer on the resulting feature matrix to reduce dimensionality in the output, and results in a $1\times1$ patch in the output which is the area of highest discrimination for the discriminator network. This allows our Generator to refine these areas in the next iteration and thus enable the overall PANDA-GAN architecture to reduce the reconstruction error substantially, gaining a better fit on the manifold over $X$. We use a Sigmoid activation function which issues a continuous probability score for a presented image on whether it is an element of the real dataset or a synthetic image produced by the Generator network. To prevent vanishing gradients, which is common with logistic functions, we use the residual network, ResNet-50 \cite{He2016} as our main backbone architecture. The pair of real, non-anomalous data examples ($x$) and the generated, synthetic examples ($x'$) from $x$ are fed into the discriminator to obtain a probability score that each of the images is an element of $X$.

Overall, the Discriminator seeks to optimise: 

\begin{equation}\label{loss_discriminant}
L_{C} = -log(C(x)) - log(1-C(x'))
\end{equation}

where $C$ represents the discriminator, or critic network. The pair ($x, x'$) obtains probability outputs $C(x)$ and $C(x')$ respectively. $L_{c}$ represents $Pr(x \in X | (x, x'))$.

\subsection{Anomaly Scoring} \label{sec:AnomalyScoring}
Anomaly scoring is the process of categorising samples as anomalous or non-anomalous based on the knowledge that the network has obtained via the gained approximation to the manifold over $X$ (normal samples) during training.

Anomalous samples will be reconstructed by the Generator model from this approximated manifold producing a normal appearing sample output upon generation. This allows us to infer a distribution of reconstruction error $N_{recon}$ over both normal and anomalous samples. When combined with the two Discriminator scores across both the input sample and the synthetically generated sample distributions $N_{C(x)}$ and $N_{C(x')}$ respectively, these offer more information during anomaly scoring. Due to the independence of these normally distributed random variables,  we can combine them into a new distribution using summation $N_{com}{\sim N(\Sigma (\mu _{i}, \sigma^{2}_{i}))} = N_{com} = N_{recon } + N_{C(x)} + N_{C(x')}$. We then normalise $N_{com}$ to values $0<x<1$ via Equation \ref{normalise} as follows: 

\begin{equation}\label{normalise}
N_{com} = \forall x \in [N_{com}], \frac{x_{i}'-x_{min}}{x_{max}-x_{min}}
\end{equation}

Once we obtain the distribution  $N_{com}$, an anomaly score is given to each sample presented to the network. A boundary between $N_{com}(x_{normal})$ and $N_{com}(x_{anomalous})$ determines whether the presented sample is anomalous on the basis of maximum likelihood.
\begin{figure*}[htb!]
\centering
\includegraphics[width=\linewidth]{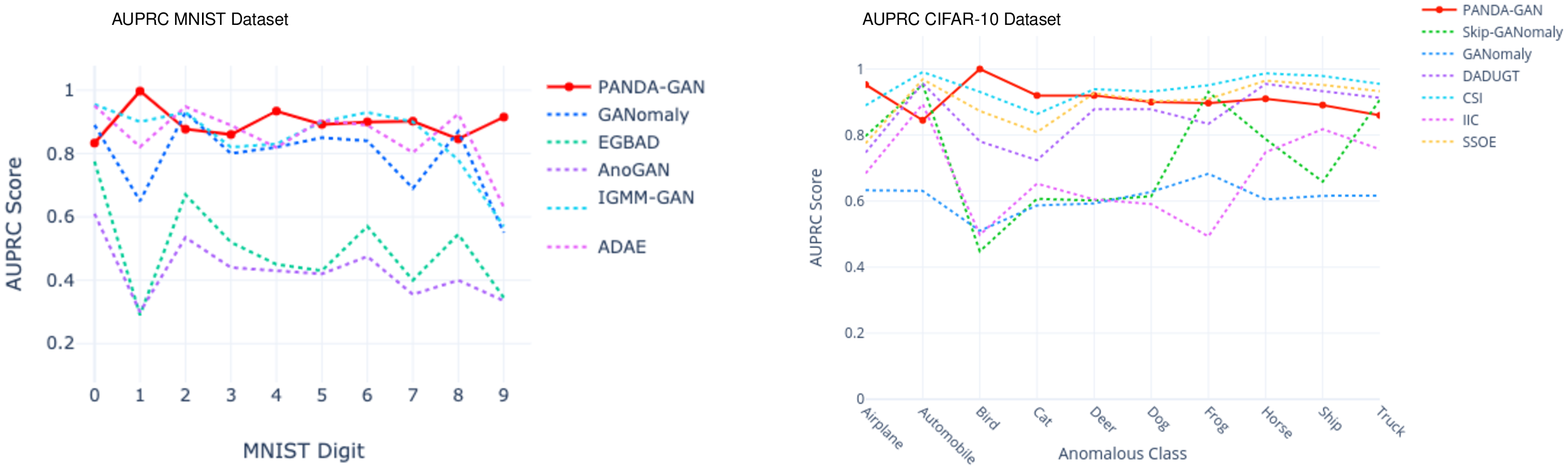}
\caption{AUPRC performance comparison of PANDA-GAN against prior state-of-the-art methods of anomaly detection across 'leave-one-out' tasks across \textbf{Left:} MNIST handwritten digit dataset  \textbf{Right:} CIFAR-10 object dataset.}
\label{fig:MNISTCIFAR}
\end{figure*}

\section{Evaluation} \label{Sec:experiment}

\subsection{Experimental Setup}
The experimental setup comprises of the following dataset configurations:

\begin{itemize}[]
\item {\textbf{Plant Village} \cite{Hughes2015}: comprises of 11 genera of agricultural plant for food with 26 diseases. The present diseases range from those which are subtle in appearance (Powdery Mildew) to diseases such as Isariopsis and Black Measles which are visually obvious.}

\item{\textbf{X-ray Security Electronics Dataset }\cite{Bhowmik2019}: comprises large consumer electronics items (e.g., laptops) with and without intra-object anomaly concealment present. Anomaly concealments consist of replica Improvised Explosive Devices (IED), metal screws, metal plates, knife blades and similar hidden inside the electronic items.}

\item{\textbf{UCSDped1} \cite{Mahadevan2010}: comprises surveillance video of pedestrian-only zone featuring pedestrians walking, running, standing, or any such similar activity. Anomalies are pedestrians riding or driving vehicles (i.e. bicycles, skateboards, scooters, vans, cars, etc) in the zone.} 

\item{\textbf{MNIST} \cite{lecun-mnisthandwrittendigit-2010}: comprises a handwritten numeric digit (0-9) leave-one-out classification task (one select digit class is anomalous, all remaining digit classes are non-anomalous) across all digits from 0 to 9.}

\item{\textbf{CIFAR-10} \cite{Krizhevsky09learningmultiple}: comprises 10 classes of tiny objects with dimension $32\times32$ such that a leave-one-out anomaly classification task (one select object class is anomalous, all remaining object classes are non-anomalous) is posed.}

\item{\textbf{MVTEC}} \cite{Bergmann2019MVTEC}: comprises of high-resolution industrial inspection data comprising of 15 object and texture classes of typical factory-line objects. Sub-classes of each object contain both visual defects of the respective object and corresponding defect-free objects.

\end{itemize}
We use the dataset split for \{\textit{train:validate:test}\} as following: \{\textit{13,593:2,589:12,661}\} for Plant Village \cite{Hughes2015}, \{\textit{229:25:125}\} for X-ray Security Electronics \cite{Bhowmik2019}. The \{\textit{train:test}\} split for MNIST and CIFAR-10 is \{80\%:20\%\} across both datasets \cite{Akcay2018, Zenati2018}. The hyper-parameters and data configurations are fine-tuned by systematic grid search in order to obtain the best results across the problems presented in this work. Pixel values in input images are normalised to a mean and a standard deviation of 0.5. All models use ADAM momentum \cite{Diedrik2015} except our Perceptual Loss model which uses Stochastic Gradient Descent (SGD) with momentum 0.9. Learning rates used: $7\times10^{-6}$ - Generator,  $1\times 10^{-5}$ - Discriminator, and  $1\times10^{-4}$ - Perceptual Loss model. Training is performed on a Nvidia Titan X GPU using a batch size of 15.

\definecolor{Gray}{gray}{0.85}

\begin{table*}[h!]
\caption{Results of models across Leaf disease \cite{Hughes2015} and X-ray Laptop Anomaly detection \cite{Bhowmik2019} image datasets as well as results across UCSDPed1 \cite{Mahadevan2010} pedestrian detection and crowd control video dataset using frame-level comparison \cite{Huang2018}.}
\begin{tabular}{|ll|l|l|l|l|l|l|l|l|l|l|}
\hline
\multirow{4}{*}{Model}  & \multirow{4}{*}{Loss} & \multicolumn{10}{c|}{Image Dataset} \\ \cline{3-12}
&  & \multicolumn{5}{c|}{\textbf{Plant Village}\cite{Hughes2015}}  & \multicolumn{5}{c|}{\textbf{Laptop X-ray}\cite{Bhowmik2019}} \\ \cline{3-12}

&  & \multicolumn{1}{c|}{AUC} & \multicolumn{1}{c|}{\begin{tabular}[c]{@{}c@{}}95\% CI \\ (AUC)\end{tabular}} & \multicolumn{1}{c|}{\begin{tabular}[c]{@{}c@{}}Average\\ Rec\_Err\end{tabular}} & \multicolumn{1}{c|}{\begin{tabular}[c]{@{}c@{}}Average\\ Adv\_Err\end{tabular}} & \multicolumn{1}{c|}{I/t(ms)} & \multicolumn{1}{c|}{AUC} & \multicolumn{1}{c|}{\begin{tabular}[c]{@{}c@{}}95\% CI \\ (AUC)\end{tabular}} & \multicolumn{1}{c|}{\begin{tabular}[c]{@{}c@{}}Average\\ Rec\_Err\end{tabular}} & \multicolumn{1}{c|}{\begin{tabular}[c]{@{}c@{}}Average\\ Adv\_Err\end{tabular}} & \multicolumn{1}{c|}{I/t(ms)} \\ \hline \hline
\rowcolor{gray!15}
\rowcolor{gray!15}
{VAE \cite{Kingma2013}}  & - & 0.65 & 0.60\textless{}x\textless{}0.70   & 0.56  & -  & 6.9 & 0.21 & 0.19\textless{}x\textless{}0.23 & 0.80 & - & 9.4 \\ 

\rowcolor{gray!0}
{AnoGAN \cite{Schlegl2017}}&-& 0.65                     & 0.65\textless{}x\textless{}0.66                                               & 0.45                                                                            & 0.88                                                                            & 7151                       & 0.41                     & 0.39\textless{}x\textless{}0.42                                               & 0.4                                                                             & 0.92                                                                            & 7223                         \\ 
\rowcolor{gray!15}
{EGBAD \cite{Zenati2018}} &-& 0.70                     & 0.65\textless{}x\textless{}0.67                                               & 0.40& 0.92                                                                            & 87                       & 0.47                     & 0.42\textless{}x\textless{}0.43                                               & 0.41                                                                            & 0.94                                                                            & 89                         \\ 
\rowcolor{gray!0}
{GANomaly \cite{Akcay2018}}                           &               - & 0.73                     & 0.68\textless{}x\textless{}0.73                                               & 0.39                                                                            & 0.75                                                                            & 28             & 0.49                     & 0.41\textless{}x\textless{}0.51                                               & 0.34                                                                            & 0.78                                                                            & 273                         \\ 
\rowcolor{gray!15}
{f-AnoGAN \cite{Schlegl2019}}                           &             -  & 0.765                    & 0.65\textless{}x\textless{}0.78                                               & 0.12                                                                            & 0.72                                                                            & 65                       & 0.50                     & 0.49\textless{}x\textless{}0.53                                               & 0.1                                                                             & 0.72                                                                            & 86                         \\ 
\rowcolor{gray!0}
{Skip-GANomaly \cite{Akcay2019}}                      &               - & 0.771                    & 0.74\textless{}x\textless{}0.77                                               & 0.13                                                                            & 0.74                                                                            & 123                      & 0.51                     & 0.48\textless{}x\textless{}0.58                                               & 0.11                                                                            & 0.68

& 112                         \\ 
\rowcolor{blue!5}
& {PWL} & \textbf{0.776}& \textbf{0.77\textless{}x\textless{}0.78}                                      & \textbf{0.012} & 0.994                                                                           & 15.2                      & 0.42                     
& 0.30\textless{}x\textless{}0.48                                               & 0.052                                                                           & 0.987                                                                           & 16.8                        \\ 
\rowcolor{blue!5}

\textbf{PANDA-GAN}& {PL$_{g}$()}  & 0.74 & 0.73\textless{}x\textless{}0.75  & 0.40   & 0.993  & 20  & 0.45 & 0.29\textless{}x\textless{}0.52  & \textbf{0.015}& 0.658  & 36\\ 
\rowcolor{blue!5}
& {PL$_{ps}$()} & 0.75 & 0.76\textless{}x\textless{}0.78  & 0.20 & 0.986 & 20.8  & \textbf{0.51}  & \textbf{0.48\textless{}x\textless{}0.55} & 0.045 & 0.782 & 30 \\ \cline{1-12} \cline{1-4}

\multirow{3}{*}{Model}  & \multirow{3}{*}{Loss} & \multicolumn{2}{c|}{Video Dataset} \\ \cline{3-4}
& & \multicolumn{2}{c|}{\bf UCSDPed1 \cite{Mahadevan2010}} \\ \cline{3-4}
& & \multicolumn{1}{c|}{AUC} & \multicolumn{1}{c|}{EER}  \\ \cline{1-4}
\rowcolor{gray!15}
SF \cite{mehran2009} &-& 0.675 & 31 \\ 
\rowcolor{gray!0}
MPPCA\cite{Kim2009}&-& 0.7696& 40 \\ 
\rowcolor{gray!15}
MDT\cite{Weixin2014}&-& 0.818& 25 \\ 
\rowcolor{gray!0}
SRC\cite{Cong2013} &-& 0.86& 19 \\
\rowcolor{gray!15}
AMDN \cite{Dan2015}&-& 0.921& 16 \\ 
\rowcolor{gray!0}
PCA-NET GMM\cite{Huang2018}&-& 0.926& 11.2 \\ 
\rowcolor{gray!15}
AED-GAN\cite{Ravanbakhsh2017}&-& \textbf{0.974} & \textbf{8} \\ 

\rowcolor{blue!5}
& PWL  & 0.945 & 35 \\ 
\rowcolor{blue!5}
\textbf{PANDA-GAN} & PL$_{g}$()& 0.95           & 75 \\ 
\rowcolor{blue!5}
& PL$_{ps}$()        & 0.93                    & 96 \\ \cline{1-4}
\end{tabular}
\label{tab:results_general}
\end{table*}

\begin{table*}[h!]
\caption{AUPRC results across MVTEC \cite{Bergmann2019MVTEC} dataset.}
\label{tab:mvtec}

\resizebox{\textwidth}{!}{
\begin{tabular}{|l|l|l|l|l|l|l|l|l|l|l|l|l|l|l|l|l|}
\hline
\multirow{2}{*}{\bf Model} & \multicolumn{16}{c|}{\textbf{Classes}} \\ \cline{2-17}
& \textbf{Bottle} & \textbf{Cable} & \textbf{Capsule} & \textbf{Carpet} & \textbf{Grid} & \textbf{Hazelnut} & \textbf{Leather} & \textbf{Metal Nut} & \textbf{Pill} & \textbf{Screw} & \textbf{Tile}  & \textbf{Toothbrush} & \textbf{Transistor} & \textbf{Wood}  & \textbf{Zipper} &\textbf{$AUC_{avg}$}\\ \hline \hline
\rowcolor{gray!15}
AnoGAN \cite{Schlegl2017}        & 0.8             & 0.477          & 0.442            & 0.337           & 0.871         & 0.259             & 0.451            & 0.284              & 0.711         & 1              & 0.401          & 0.439               & 0.692               & 0.567          & 0.715         &0.563  \\ 
GANomaly\cite{Akcay2018}      & 0.794           & \textbf{0.711} & 0.721            & 0.821           & 0.743         & 0.874             & 0.808            & 0.694              & 0.671         & 1              & 0.72           & 0.7                 & 0.808               & 0.92           & 0.744          &0.782 \\ 
\rowcolor{gray!15}
Skip-GANomaly\cite{Akcay2019}  & 0.937           & 0.674          & 0.718            & 0.795           & 0.657         & 0.906             & 0.908            & 0.79               & 0.758         & 1              & 0.85           & 0.689               & 0.814               & 0.919          & 0.663       &0.805    \\ 

DA-GAN \cite{Tang2020}        & \textbf{0.983}  & 0.665          & 0.687            & 0.903           & 0.867         & \textbf{1}        & \textbf{0.944}   & \textbf{0.815}     & 0.768         & 1              & 0.961          & \textbf{0.95}                & 0.794               & \textbf{0.979} & \textbf{0.781} & \textbf{0.873} \\ 
\rowcolor{gray!15}
U-Net \cite{Ronneberger2015}        & 0.863           & 0.636          & 0.673            & 0.774           & 0.857         & 0.996             & 0.87             & 0.676              & 0.78          & 1              & \textbf{0.964} & 0.811               & 0.674               & 0.958          & 0.75         &	0.819   \\
\rowcolor{blue!8}
\textbf{PANDA-GAN}     & 0.826           & 0.68           & \textbf{0.98}    & \textbf{0.95}   & \textbf{0.95} & 0.922             & 0.75             & 0.79               & \textbf{0.95} & \textbf{1}     & 0.85           & 0.66                & \textbf{0.9}        & 0.68           & 0.62       &0.834     \\ \hline
\end{tabular}
}

\end{table*}

\subsection{Results and Discussion}
\label{Sec:results}

\begin{table*}[h!]
\caption{Ablation Study of PANDA-GAN across Plant Village \cite{Hughes2015} and Laptop Anomaly \cite{Bhowmik2019}. }
\label{table:ablation}
\resizebox{\textwidth}{!}{

\begin{tabular}{|l|c|c|c|c|c|c|c|c|c|c|c|c|}
\hline
\multirow{4}{*}{\bf Model} & \multicolumn{12}{c|}{\textbf{Dataset}} \\ \cline{2-13} 
& \multicolumn{6}{c|}{\textbf{Plant Village}} & \multicolumn{6}{c|}{\textbf{Laptop Anomaly}} \\ \cline{2-13} 
& \multicolumn{3}{c|}{\textbf{Loss}} & \multicolumn{3}{c|}{\textbf{Network Architecture}}   & \multicolumn{3}{c|}{\textbf{Loss}} & \multicolumn{3}{c|}{\textbf{Network Architecture}}   \\ \cline{2-13}
& PWL       & PL(g)     & PL(ps)     & $E^{0}_{high} \cap D^{0}_{high}$ & $E^{1}_{high}$ & $E^{1}_{low}$ & PWL       & PL(g)     & PL(ps)     & $E^{0}_{high} \cap D^{0}_{high}$ & $E^{1}_{high}$ & $E^{1}_{low}$ \\ \hline \hline
\multirow{6}{*}{\textbf{PANDA-GAN}} & 0.75      & 0.754     & 0.762      & \xmark                           & -              & \xmark        & 0.38      & 0.42      & 0.434      & \xmark                           & -              & \xmark        \\  
 & 0.752     & 0.74      & 0.746      & \xmark & -  & \cmark        & 0.419     & 0.442     & 0.446      & \xmark & -  & \cmark        \\  
& 0.751     & 0.738     & 0.744      & \cmark                           & \xmark         & \xmark        & 0.464     & 0.468     & 0.476      & \cmark                           & \xmark         & \xmark        \\  
& 0.764     & 0.751     & 0.762       & \cmark                           & \xmark         & \cmark        & 0.462     & 0.442     & 0.43       & \cmark                           & \xmark         & \cmark        \\  
& 0.771     & \textbf{0.764}     & \textbf{0.775}      & \cmark                           & \cmark         & \xmark        & \textbf{0.496}     & \textbf{0.515}     & 0.504      & \cmark                           & \cmark         & \xmark        \\  
& \textbf{0.776}     & 0.741     & 0.769     & \cmark                           & \cmark         & \cmark        & 0.42      & 0.451     & \textbf{0.512}       & \cmark                           & \cmark         & \cmark        \\ 
\rowcolor{gray!15}
\shortstack[l]{\bf PANDA-GAN\\ \bf DCGAN Discriminator}
& 0.769     & 0.751     & 0.74       & \cmark                           & \cmark         & \cmark        & 0.413     & 0.424     & 0.469      & \cmark                           & \cmark         & \cmark        \\ \hline
\end{tabular}
}
\end{table*}

The AUPRC statistical score across the classical `leave-one-out' anomaly detection tasks (MNIST / CIFAR-10) are outlined in Figure \ref{fig:MNISTCIFAR} where it can be observed that our approach (PANDA-GAN) outperforms prior state-of-the-art approaches on these seminal, albeit unrealistic benchmark tasks.

As is evident from Figure \ref{fig:MNISTCIFAR}, further direct comparison of model performance solely across these `leave-one-out' MNIST / CIFAR-10 based anomaly detection tasks is becoming decreasingly informative due to potential performance saturation among competing approaches. Across the MNIST task (left) it can be seen that our method obtains state-of-the-art results across 40\% of classes and obtains close performance to the other prior methods while exhibiting uniform performance across all classes. Most noticeably is the result across the digit 9 whereby PANDA-GAN is over 0.2 AUPRC higher than prior methods. Across CIFAR-10 (right), our method obtains state-of-the-art in 30\% of classes and matches closely with other such methods (DADUGT \cite{Golan2018}, CSI\cite{Tack2020}, and SSOE \cite{Hendrycks2019}) while also obtaining close to uniform performance across all classes.

By contrast, Table \ref{tab:results_general} outlines quantitative results across the challenging real-world benchmark datasets of Plant Village \cite{Hughes2015}, Laptop X-ray \cite{Bhowmik2019} and UCSDPed1 \cite{Mahadevan2010} providing numerous statistical comparatives including Area Under Curve (AUC), the 95\% confidence interval of the AUC, inference time (I/t, ms) per image.  Datasets \cite{Hughes2015, Bhowmik2019} feature particularly subtle anomalies by nature and as such pose as challenging tasks for semi-supervised anomaly detection models. 


PANDA-GAN obtains the highest AUC value across both image based datasets (Plant Village: 0.776- using Pixel-Wise Loss (PWL); Laptop X-ray: 0.51- using Problem Specific Perceptual Loss(PL$_{ps}$)) in comparison to leading state-of-the-art methods \cite{Simonyan15, Kingma2013, Schlegl2017, Schlegl2019, Zenati2018, Akcay2018, Akcay2019} (Table \ref{tab:results_general}). Over multiple evaluations, PANDA-GAN also obtains tighter confidence-intervals compared to prior semi-supervised work illustrating our PANDA method can produce more stable and reliable results across the same dataset while other such approaches can suffer from sporadic performance at inference (Table \ref{tab:results_general}). Observing the I/t(ms), the PANDA method is significantly faster than prior methods.


In Table \ref{tab:mvtec}, the quantitative AUC results across the MVTEC dataset can be observed. This is a challenging dataset due to the large variation in appearance of anomalies present in textures and objects. Some objects (carpet, hazelnut, screw) exhibit visually distinct and obvious anomalies, but other objects (cable, metal nut, toothbrush) feature subtle anomalies which are hard to detect. This is reflected in the results of various methods across this dataset with PANDA-GAN obtaining superior AUC performance across 6 classes. 

Our ablation study (Table \ref{table:ablation}) produces evaluation over individual components to our novel architecture with respect to variations in both loss function, our network architecture components ($E^{0}_{high} \cap D^{0}_{high}$, $E^{1}_{high}$, $E^{1}_{low}$) and our choice of discriminator architecture across two of the more challenging real-world anomaly detection task datasets. For comparison we include the DCGAN \cite{Radford2016} discriminator architecture from GANomaly / Skip-GANomaly \cite{Akcay2018, Akcay2019}, which is the next best performing approach in terms of AUC across the same datasets (Table \ref{tab:results_general} to compare against our FGVC-based discriminator architecture choice.

From the results of Table \ref{table:ablation}, it can be observed that synergy exists between components of our generator network obtaining the highest AUC value only when all three of our novel components are activated. Generally we see that the more components we activate in our architecture, the better the performance obtained during our ablation study. Overall, the problem-specific perceptual loss ($PL_{ps}$) performs better across the Laptop X-ray dataset by a clear margin from the other loss functions tested against. Across the Plant Village dataset, there is negligible difference between the pixel-wise loss and the problem specific perceptual loss. Both performed almost identically and gained a clear advantage over the general perceptual loss function ($PL_{g}$).

\begin{figure}[htb!]
\centering
\includegraphics[width=160pt]{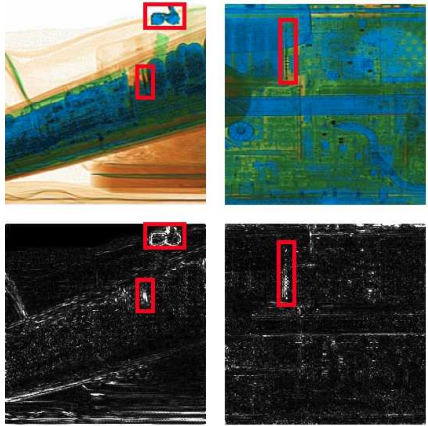}
\caption{\textbf{Top:} X-ray images of large electronic items (Anomalous region ground truth outlined with red bounding box). \textbf{Bottom:} Non-threshold instance segmentation masks produced by PANDA-GAN outlining anomalous artifacts within X-ray scans.}
\label{fig:mvtec_detections}
\end{figure}

\begin{figure}[htb!]
\centering
\includegraphics[width=230pt]{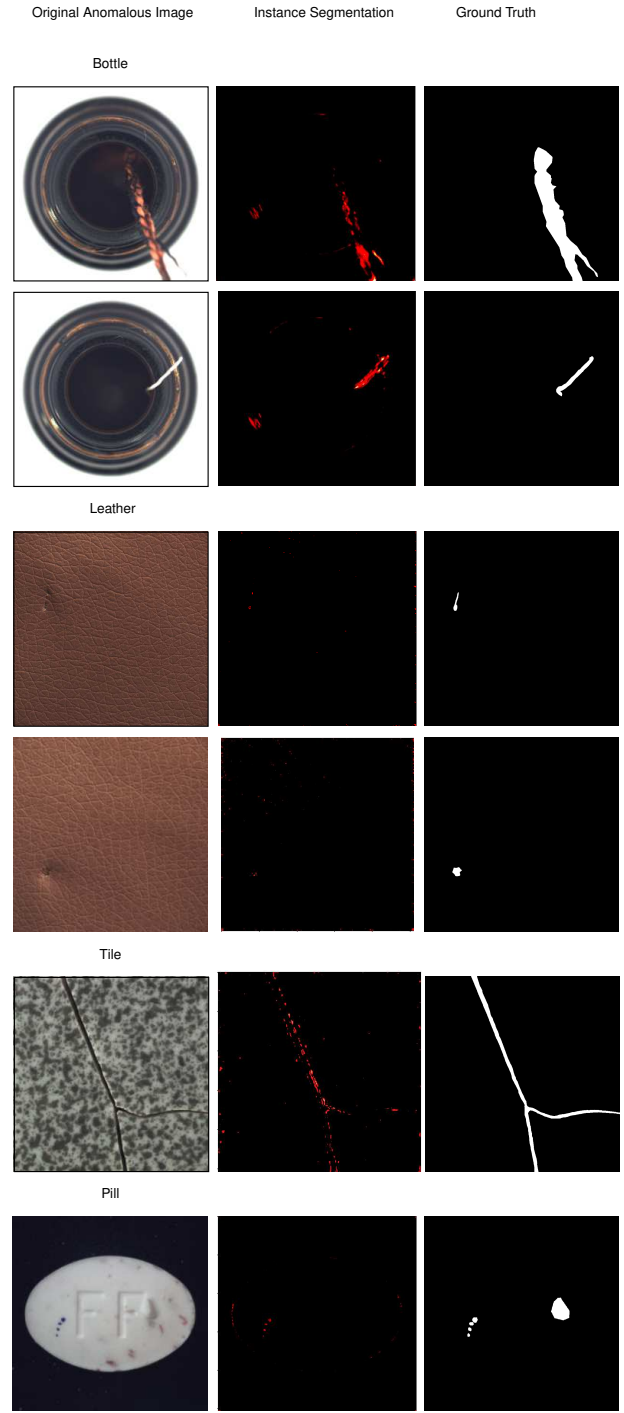}
\caption{Anomalous regions obtained from PANDA-GAN across the MVTEC dataset together with the original image and the ground truth for the anomalous region.}
\label{fig:mvtec_detections}
\end{figure}

\section{Conclusion}

We propose the semi-supervised method of Perceptually Aware Neural Detection of Anomalies (PANDA-GAN), a Variational Autoencoder Generative Adversarial Network (VAE-GAN) based architecture. PANDA-GAN includes three novel proposals: a Fine-Grained Visual Categorisation Discriminator Network (FGVC) to ease the problem of low-inter class variance present in anomaly detection problems, a dual-latent space implementation that carries higher-level features in given images forward in the architecture and the use of a perceptual loss function to compute feature error which enables higher aptitude at detecting anomalies within the given problems presented in this work.

Our PANDA-GAN architecture obtains superior results across: (CIFAR-10, AUPRC$_{avg}$: 0.91; MNIST, AUPRC$_{avg}$: 0.90) and challenging real-world datasets (Plant Leaf Disease, AUC: 0.776, Threat Item X-ray, AUC: 0.51, MVTEC, AUC$_{avg}$: 0.83), in addition to video based frame-level anomaly detection (UCSDPed1, AUC: 0.95). Outperforming prior work by \cite{Schlegl2017, Zenati2018, Akcay2018, Schlegl2019, Akcay2019, mehran2009, Kim2009, Mahadevan2010, Cong2013, Dan2015, Huang2018, Smolyak2020, Vu2019, Golan2018}.

{\small
\bibliographystyle{IEEEbib}
{\footnotesize
\bibliography{biblio}}
}

\end{document}